%% file: colm2026_conference.tex
\documentclass{article} % For LaTeX2e
\usepackage[preprint]{Template-2026/colm2026_conference}

\usepackage{microtype}
\usepackage{hyperref}
\usepackage{url}
\usepackage{booktabs}

\usepackage{lineno}

\usepackage{amsmath}
\usepackage{bm}
\usepackage{graphicx}
\newcommand{\Wt}{\tilde{\bm{W}}}
\newcommand{\Rt}{\tilde{\bm{R}}}
\newcommand{\qt}{\tilde{\bm{q}}}
\newcommand{\kt}{\tilde{\bm{k}}}

\newcommand{\xt}{\tilde{\bm{x}}}
\newcommand{\W}{{\bm{W}}}

\newcommand{\q}{{\bm{q}}}
\newcommand{\x}{{\bm{x}}}
\newcommand{\p}{{\bm{p}}}
\renewcommand{\b}{{\bm{b}}}
\newcommand{\norm}[1]{\left\lVert #1 \right\rVert}

\definecolor{darkblue}{rgb}{0, 0, 0.5}
\hypersetup{colorlinks=true, citecolor=darkblue, linkcolor=darkblue, urlcolor=darkblue}

\title{CRoPE: \\Efficient Parametrization of Rotary Positional Embedding}

\author{%
  Beicheng Lou\thanks{Equal contribution} \\
  Stanford University \\
  Stanford, CA 94305 \\
  \texttt{beichenglou@stanford.edu} \\
  \And
  Zifei Xu\footnotemark[1] \\
  d-Matrix \\
  Santa Clara, CA 95054 \\
  \texttt{xuzifei@d-matrix.ai} \\
  \And
  Vivian W. H. Wong\\
  University of Florida\\
  Gainesville, FL 32611 \\
  \texttt{vivian.wong@ufl.edu} \\
}

\begin{document}

\ifcolmsubmission
\linenumbers
\fi

\maketitle

\begin{abstract}
  Rotary positional embedding has become the state-of-the-art approach to encode position information in transformer-based models. While it is often succinctly expressed in complex linear algebra, we note that the actual implementation of $Q/K/V$-projections is not equivalent to a complex linear transformation. We argue that complex linear transformation is a more natural parametrization and saves near 50\% parameters within the attention block. We show empirically that removing such redundancy has negligible impact on the model performance. Our modification achieves more efficient parameter usage, as well as a cleaner interpretation of the representation space. 
\end{abstract}

\section{Introduction}

Transformer has become the state-of-the-art architecture for large language and time series modeling tasks \citep{vaswani2023attentionneed,openai2024gpt4technicalreport,deepseek-ai_deepseek-v3_2025,team_gemini_2025}. At its core, it uses the attention mechanism to route information through the most relevant paths and allow different parts of the input to interact synergistically. Unlike recurrent neural networks, the attention mechanism does not inherently encode token order, so positional information must be explicitly injected.

Positional embedding is crucial for transformer since its birth \citep{dufter_position_2021}. The choice of positional embedding scheme also significantly affects training dynamics \citep{kazemnejad_impact_2023}. As models scale and generalize to longer contexts, careful treatment of positional encoding becomes increasingly important \citep{peng_yarn_2023}.

However, positional embedding schemes have never been perfect \citep{chen_simple_2021,ke_rethinking_2021}. Early absolute embeddings made it hard for models to disentangle position from semantic content \citep{vaswani2023attentionneed,devlin_bert_2019}. Relative embeddings mitigated this but required extra parameters \citep{dai_transformer-xl_2019,raffel_exploring_2023,he_deberta_2021}. Rotary positional embedding (RoPE) removes the explicit parameterization, yet implicitly still reserves half of the embedding space for positional information \citep{su_roformer_2023}. The search for more efficient encoding schemes is not finished.

In this paper, we revisit a complex-valued formulation of RoPE that appears equivalent to the original work at first glance \citep{wang_encoding_2020}, but with a fundamental difference in the function space. From this perspective, we argue that a more natural parameterization of attention would require 50\% fewer parameters within the $Q/K/V$-projections, with minimal performance loss. The ratio of saved parameters drops below 50\% when counting other components such as feedforward and embedding layers, but remains significant and can be invested elsewhere.

\section{Background and related work}
\subsection{Absolute and relative positional embedding}
Essentially, positional embedding maps $i=1,2,...,L$ input embeddings to $L$ output embeddings, where each output embedding attends to preceding input embeddings with different weighting.

For any two input embeddings $\bm{x}_m, \bm{x}_n \in\mathbb{R}^D,\in\mathrm{R}^D,\; m,n\in \mathbb{Z}^+$, we want
\begin{equation}
    attn(m\rightarrow n) = f(\bm{x}_m, \bm{x}_n, m, n)
    \label{eq:attn-general}
\end{equation}
so that the attention weights can be calculated from 
\begin{equation}
    \bm{a}_{m,n} = \frac{\exp{attn(m\rightarrow n)}}{\sum_j \exp attn(m\rightarrow j)}
    \label{eq:attn-weights}
\end{equation}
and the output embedding is a weighted average according to the attention weights:
\begin{equation}
    \bm{o}_{m} = \sum_{j=1}^L \bm{a}_{m,j}\bm{v}_j
    \label{eq:attn-out}
\end{equation}
where $\bm{v}_j$ is embeddings derived from $\bm{x}_j$.

In absolute positional embedding\citep{vaswani2023attentionneed,devlin_bert_2019}, one simply chooses:
\begin{equation}
    f(\bm{x}_m, \bm{x}_n, m, n) = (\bm{x}_m+\bm{p}_m)^T\bm{W}_q^T \bm{W}_k (\bm{x}_n+\bm{p}_n)
    \label{eq:pe-abs}
\end{equation}
There could be other variations, but the impact is less significant.

In relative positional embedding\citep{dai_transformer-xl_2019,raffel_exploring_2023,he_deberta_2021}, one forces the function to be only a function of $m-n$. One common choice is:
\begin{equation}
    f(\bm{x}_m, \bm{x}_n, m-n) = (\bm{x}_m+{\bm{p}}_{m-n})^T\bm{W}_q^T \bm{W}_k (\bm{x}_n+{\bm{p}}_{m-n})
    \label{eq:pe-rel}
\end{equation}

\subsection{Rotary Positional Embedding}
In RoPE\citep{su_roformer_2023}, one has a rotation matrix that performs position-dependent rotations to each 2-by-2 subspace in the following form:
{\small
\begin{equation}
\mathbf{R}_m = 
\begin{pmatrix}
\cos(m\theta_1) & -\sin(m\theta_1) & 0 & 0 & \cdots & 0 & 0 \\
\sin(m\theta_1) &  \cos(m\theta_1) & 0 & 0 & \cdots & 0 & 0 \\
0 & 0 & \cos(m\theta_2) & -\sin(m\theta_2) & \cdots & 0 & 0 \\
0 & 0 & \sin(m\theta_2) &  \cos(m\theta_2) & \cdots & 0 & 0 \\
\vdots & \vdots & \vdots & \vdots & \ddots & \vdots & \vdots \\
0 & 0 & 0 & 0 & \cdots & \cos(m\theta_L) & -\sin(m\theta_L) \\
0 & 0 & 0 & 0 & \cdots & \sin(m\theta_L) &  \cos(m\theta_L)
\end{pmatrix}
\end{equation}
}
and Eq.~\ref{eq:attn-general} simply becomes:
\begin{align}
    f(\bm{x}_m, \bm{x}_n, m-n) &= \bm{x}_m^T\bm{W}_q^T\mathbf{R}_m^T \mathbf{R}_n \bm{W}_k \bm{x}_n\notag\\
    &= \bm{x}_m^T\bm{W}_q^T\mathbf{R}_{m-n}^T \bm{W}_k \bm{x}_n
    \label{eq:pe-rope}
\end{align}
The remaining procedure from Eq.~\ref{eq:attn-weights} onwards is the same as in the original absolute positional embedding case.

\subsection{Related work}
%?! relate to weight tying
Since our work described a method that achieved similar performance with significantly fewer parameters, it could be reminiscent of other pruning or compression work.
Typical pruning methods rely on the information in hidden states or hessian information \citep{sparsegpt,wanda}.
There is a detailed tradeoff between the number of parameters pruned and the performance decay, and the optimal choice is highly dependent on the specific utility function. 
In our work, the parameter efficiency is obtained through an architectural inductive bias, which is both convenient and safe from noise in data.
Architecture search could also lower the number of parameters \citep{real_automl-zero_2020,liu_darts_2019}, but it requires huge effort and is subject to noise in data. In the end, it may not arrive at the same architectural inductive bias we manually introduced.
% The scale of parameter saving (nearly half) is similar to that of quantization . We note that our method is completely different. 
Since it works on the architecture level, it is fully compatible with any additional memory saving optimization, e.g. quantization \citep{lin2024awqactivationawareweightquantization,frantar_sparsegpt_2023,resq}. Similarly, one could always perform pruning and compression starting from our parametrization.

There are other approaches to improve parameter efficiency at run time instead of in architecture design. In mixture-of-expert architectures \citep{shazeer_outrageously_2017, lou_meta-learning_2021,fedus_switch_2022, gale_megablocks_2022}, some blocks of the model can be entirely skipped during run time. In contrast, our modification applies to a more minuscule scale and is fully compatible with MoE.
It is also possible to skip some part of network and therefore run through fewer parameters through early exit \citep{schuster_confident_2022}.
Similarly, it can also be applied on top of our modifications.

Various other papers aim to modify RoPE in different scenarios.
YARN \citep{peng_yarn_2023} discussed how to finetune an existing model to work on longer sequence lengths than what it was trained for. Our modification introduces a different parametrization with simpler function space and could potentially make the finetuning dynamics better. Our modification can also be viewed as intra-matrix weight tying in a structured way that respects Cauchy-Riemann symmetry \citep{trabelsi2018deep}.

% Pruning is a common approach to reduce parameters within a model that can achieve minimal accuracy degradation without the need of finetuning. ~\citep{wanda} prunes weights that are multiplied with small activations, ~\citep{sparsegpt} uses second order Hessian information to prune weights 

\input{theory}

\input{methods}
\input{results}

\section{Conclusion}
In conclusion, we have shown that rewriting RoPE in complex form naturally leads to CRoPE, in which the $Q/K/V$ projections are implemented as complex-linear transformations, reducing the number of attention parameters by 25\% in \texttt{qk}, 37.5\% in \texttt{qkv}, and up to 50\% in \texttt{all}. We show empirically that these savings come with little to no noticeable degradation in model quality: training remains stable, CRoPE consistently outperforms parameter-matched dense \texttt{half\_rope} baselines, and downstream zero-shot performance remains competitive with standard RoPE. Our study therefore provides a new perspective on implementing weights in the complex form.
%%% comment out for arxiv;
% \input{limitations}

\bibliographystyle{Template-2026/colm2026_conference}  
\bibliography{references}

\appendix
\section{Appendix}
\subsection{Details for Illustrative Examples}
In the example of simple token comparison, one uses the dimensions with $l>l_t$ for the token embedding. For Eq.~\ref{eq:token-rope1}, the token embedding is simply:
\begin{equation}
    \bm{x} = 
    \begin{bmatrix}
    0 \\
    \vdots \\
    0 \\
    x_{l_t}\\
    \vdots \\
    x_{D}
    \end{bmatrix}
\end{equation}
where $\W_{t\in\{q,k\}}=\mathbf{I}$. Namely, the model simply needs to learn to store token-specific information into low-frequency bases that are almost independent of position.

In the example of token-dependent position comparison, Eq.~\ref{eq:delta-rope} results from the following choice of projection matrix:
{\small
\begin{align}
    \Wt_q = \begin{bmatrix}
        e^{-i\theta_1} & e^{-2i\theta_1} & \dots & e^{-i\theta_1} & e^{-2i\theta_1}\\
        e^{-i\theta_2} & e^{-2i\theta_2} & \dots & e^{-i\theta_2} & e^{-2i\theta_2}\\
        \vdots & \vdots & \ddots & \vdots & \vdots\\
        e^{-i\theta_{D/2}} & e^{-2i\theta_{D/2}} & \dots & e^{-i\theta_{D/2}} & e^{-2i\theta_{D/2}}
    \end{bmatrix},\, 
    \xt^{(n)}_i = \begin{bmatrix}
        a_1\\0\\a_2\\0\\ \vdots \\a_{D/4} \\ 0
    \end{bmatrix}, \,
    \xt^{(nn)}_i = \begin{bmatrix}
        0\\a'_1\\0\\a'_2\\0\\ \vdots \\ 0 \\a'_{D/4}
    \end{bmatrix}
\end{align}
}
Here $\qt_i = \Wt_q \xt_i$, where $\xt_i$ is the embedding of the $i$-th token in complex form. $a_t$ and $a_t'$ are the degrees of freedom to encode the token information. As long as $\sum_t a_t = \sum_t a'_t$, we can get Eq.~\ref{eq:delta-rope} to hold.

\subsection{Details for Training Settings}
We apply a 50-step linear warmup, then cosine decay from $2 \times 10^{-3}$ to $4 \times 10^{-4}$. The batch size is 64 sequences, and each run is trained for 30{,}000 steps. Training uses \texttt{bfloat16} autocast with distributed data-parallel execution launched via \texttt{torchrun}.

\end{document}

%% file: theory.tex
\section{Complex Rotary Positional Embedding (CRoPE)}
\label{theory}
\subsection{Origin}
Since RoPE involves rotation in various 2-dimensional subspaces, it can be easily cast to a complex form as below:
% \begin{align}
% \bm{q}_n = 
% \begin{bmatrix}
% q_{n,1} \\
% q_{n,2} \\
% \vdots \\
% q_{n,D}
% \end{bmatrix} \quad &\rightarrow \quad
% \tilde{\bm{q}}_n =
% \begin{bmatrix}
% q_{n,1} + q_{n,2} \mathrm{i}\\
% q_{n,3} + q_{n,4} \mathrm{i}\\
% \vdots \\
% q_{n,D-1} + q_{n,D} \mathrm{i}
% \end{bmatrix}
% \label{eq:q-to-complex}\\
% \bm{k}_n = 
% \begin{bmatrix}
% k_{n,1} \\
% k_{n,2} \\
% \vdots \\
% k_{n,D}
% \end{bmatrix} \quad &\rightarrow \quad
% \tilde{\bm{k}}_n =
% \begin{bmatrix}
% k_{n,1} + k_{n,2} \mathrm{i}\\
% k_{n,3} + k_{n,4} \mathrm{i}\\
% \vdots \\
% k_{n,D-1} + k_{n,D} \mathrm{i}
% \end{bmatrix}
% \label{eq:k-to-complex}
% \end{align}
{\small
\begin{align}
\bm{q}_n = 
\begin{bmatrix}
q_{n,1} \\
q_{n,2} \\
\vdots \\
q_{n,D}
\end{bmatrix} \;\rightarrow \;
\tilde{\bm{q}}_n =
\begin{bmatrix}
q_{n,1} + q_{n,2} \mathrm{i}\\
q_{n,3} + q_{n,4} \mathrm{i}\\
\vdots \\
q_{n,D-1} + q_{n,D} \mathrm{i}
\end{bmatrix}
,\quad
\bm{k}_n = 
\begin{bmatrix}
k_{n,1} \\
k_{n,2} \\
\vdots \\
k_{n,D}
\end{bmatrix} \;\rightarrow \;
\tilde{\bm{k}}_n =
\begin{bmatrix}
k_{n,1} + k_{n,2} \mathrm{i}\\
k_{n,3} + k_{n,4} \mathrm{i}\\
\vdots \\
k_{n,D-1} + k_{n,D} \mathrm{i}
\end{bmatrix}
\label{eq:k-to-complex}
\end{align}
}

One can rewrite the rotation matrix as a diagonal matrix that applies position-dependent phase:
{\small
\begin{equation}
\tilde{\mathbf{R}}_m = 
\begin{pmatrix}
e^{im\theta_1}& 0 & \cdots & 0 \\
0 & e^{im\theta_2} & \cdots & 0 \\
\vdots & \vdots & \ddots & \vdots \\
0 & 0 & \cdots & e^{im\theta_{D/2}}
\end{pmatrix}
\end{equation}
}

Note that 
\begin{align}
     \bm{q}_m^T \bm{k}_n = \operatorname{Re}[\tilde{\bm{q}}_m^{\dagger} \tilde{\bm{k}}_n]
\end{align}
and therefore Eq.~\ref{eq:attn-general} now becomes
\begin{align}
    f(\bm{x}_m, \bm{x}_n, m-n) &= \operatorname{Re}[\tilde{\bm{q}}_m^{\dagger} \tilde{\mathbf{R}}^*_{m-n} \tilde{\bm{k}}_n]
    \label{eq:pe-crope0}
\end{align}

Note that Eq.~\ref{eq:pe-crope0} is exactly equivalent to 
Eq.~\ref{eq:pe-rope}.

One might be tempted to write the input embedding $\bm{x}_m, \bm{x}_n$ in complex forms too and have
\begin{align}
    \tilde{f}(\bm{x}_m, \bm{x}_n, m-n) &= \operatorname{Re}[\xt_m^\dagger\Wt_q^\dagger \Rt^*_{m-n} \Wt_k \xt_n]
    \label{eq:pe-crope}
\end{align}

However, Eq.~\ref{eq:pe-crope} is no longer equivalent to Eq.~\ref{eq:pe-crope0}.

\begin{figure}[h]
    \centering
    \includegraphics[width=0.8\textwidth]{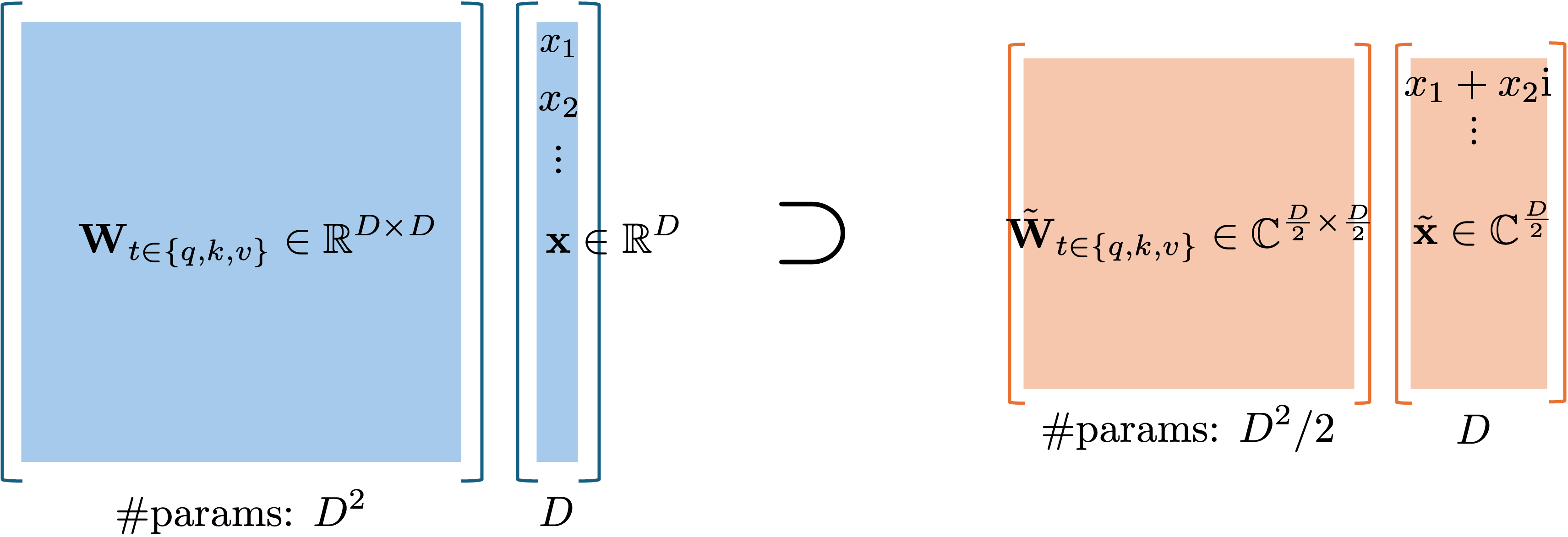}
    \caption{Main difference between RoPE (left) and CRoPE (right): the latter has a smaller function space despite having same number of parameters in the embedding}
    \label{fig:nparams}
\end{figure}

To see how the complex form of Eq.~\ref{eq:pe-crope} is not equivalent to the original RoPE, unlike the case for Eq.~\ref{eq:pe-crope0}, one simply needs to count the degrees of freedom, i.e. the number of parameters, as illustrated in Fig.~\ref{fig:nparams}.

Namely, if one casts RoPE to a complex form, the formulation naturally invites the use of a complex $\Wt$ matrix, which only has 50\% of the parameters compared to the original $\W$ matrix.

\subsection{Detailed look into function space}
While CRoPE arises naturally under the interpretation of the embedding as complex numbers, its function space is only half the size of the original RoPE. To see which half is missing, we consider the case of $D=2$.

\begin{figure}[h]
    \centering
    \includegraphics[width=0.92\textwidth]{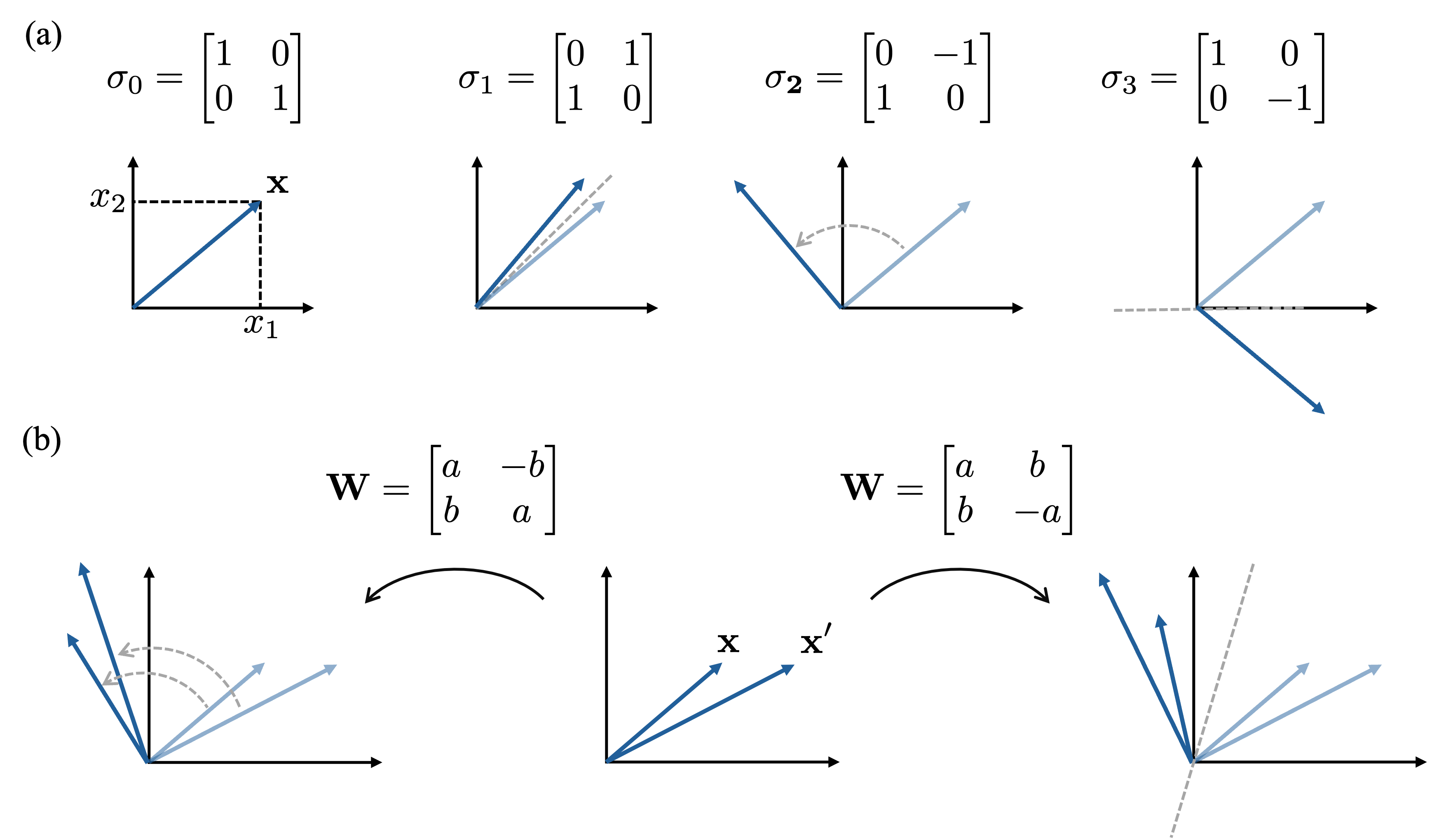}
    \caption{Main difference between RoPE (left) and CRoPE (right): the latter has a smaller function space despite having same number of parameters in the embedding}
    \label{fig:fspace}
\end{figure}

As illustrated in Fig.~\ref{fig:fspace}(a), the 2-by-2 matrix can be decomposed into four bases: $\bm{\sigma}_0,\bm{\sigma}_1,\bm{\sigma}_2,\bm{\sigma}_3$. 
Each of them can have a distinctive geometric interpretation.
$\bm{\sigma}_0$ is the identity mapping. $\bm{\sigma}_1$ reflects the vector about the line at 45 degrees. $\bm{\sigma}_2$ rotates the vector in the plane by 90 degrees. $\bm{\sigma}_3$ reflects the vector about the x-axis. 
The function space of CRoPE only utilizes $\bm{\sigma}_0,\bm{\sigma}_2$. As illustrated in Fig.~\ref{fig:fspace}(b) on the left, this corresponds to rotation in addition to length scaling. Meanwhile, CRoPE is missing out on reflections, as illustrated in Fig.~\ref{fig:fspace}(b) on the right.
Both transformations on the left and on the right are capable of mapping a single vector to anywhere in the 2D plane. Their expressive power differs when multiple vectors are considered at the same time. Having the capability of reflection for sure adds to the expressivity to the model. Whether that expressivity is worth the parameters is a different story, which depends on interpretation, tasks and various other factors. 
For example, there used to be works on adding parameters for activation, which also contribute to better model expressivity. However, it was later realized that the benefits were marginal and the state-of-the-art architectures today no longer have activations parametrized.

\section{Illustrative Example}

While reflections are definitely a useful thing to have and contribute to model expressivity, the question here is whether it is necessary to have, because the parameters we saved here can be potentially used for more dimensions to apply rotations in.

To answer that, we take a step back and review what each layer of attention is able to achieve. In these simple examples, we can analytically work out a functional solution for all the model parameters. As the dimension approaches infinity, these analytical solutions can be near perfect.

\subsection{Simple token comparison}
One basic mechanisms is token comparison, e.g. attending to similar tokens.

In absolute positional encoding, this can be achieved by increasing the scale of embedding weights. Namely, we can have $\norm{\x_i}\gg\norm{\p_i}$ in Eq.~\ref{eq:pe-abs} and $\W_{t\in\{q,k\}}=\mathbf{I}$. Therefore,
\begin{align}
    f(\bm{x}_m, \bm{x}_n, m, n) &= (\bm{x}_m+\bm{p}_m)^T\bm{W}_q^T \bm{W}_k (\bm{x}_n+\bm{p}_n)\\
    &\approx \bm{x}_m^T\bm{x}_n
    \label{eq:token-abs1}
\end{align}
which takes larger value when the tokens $\bm{x}_m$ and $\bm{x}_n$ are similar.

In RoPE, this can be achieved by encoding the token embedding in the dimensions with longest wavelengths. Namely, in Eq.~\ref{eq:pe-rope}, say $\theta_1, \theta_2, ...,  \theta_L$ are arranged in ascending order. For a given window length $w$, there exists a threshold length $l_t$ such that $\frac{1}{\theta_l}\gg w$ when $l\ge l_t$.
% Then one simply uses the dimensions with $l>l_t$ for the token embedding:
% \begin{equation}
%     \bm{x} = 
%     \begin{bmatrix}
%     0 \\
%     \vdots \\
%     0 \\
%     x_{l_t}\\
%     \vdots \\
%     x_{D}
%     \end{bmatrix}
% \end{equation}
% and uses $\W_{t\in\{q,k\}}=\mathbf{I}$. 
One can simply use the dimensions with $l>l_t$ for the token embedding and uses $\W_{t\in\{q,k\}}=\mathbf{I}$.
Then Eq.~\ref{eq:pe-rope} becomes:
\begin{align}
    f(\bm{x}_m, \bm{x}_n, m-n) &= \bm{x}_m^T\bm{W}_q^T\mathbf{R}_{m-n}^T \bm{W}_k \bm{x}_n\\
    &\approx \bm{x}_m^T\bm{x}_n
    \label{eq:token-rope1}
\end{align}

In CRoPE, this mechanism can be achieved in the same way.

\subsection{Simple position comparison}
Another fundamental mechanisms is position comparison, e.g. attending to near positions.

In absolute positional encoding, this can be achieved by reducing the scale of embedding weights. Namely, we can have $\norm{\x_i}\ll\norm{\p_i}$ in Eq.~\ref{eq:pe-abs} and $\W_{t\in\{q,k\}}=\mathbf{I}$. Therefore,
\begin{align}
    f(\bm{x}_m, \bm{x}_n, m, n) &= (\bm{x}_m+\bm{p}_m)^T\bm{W}_q^T \bm{W}_k (\bm{x}_n+\bm{p}_n)\\
    &\approx \bm{p}_m^T\bm{p}_n
    \label{eq:token-abs2}
\end{align}
which takes larger value when the position encodings $\bm{p}_m$ and $\bm{p}_n$ are similar, i.e. when $m$ and $n$ are close. To the limit of large size of dimensions $D$, 
\begin{equation}
    \lim_{D\to \infty} \bm{p}_m^T\bm{p}_n=\delta(m-n)
\end{equation} 

In RoPE, this can be achieved by setting the token embeddings to constant. This is easy when the embedding mapping involves a bias vector $\x=\W_e \x_{prev} + \b_e$, where we can set $\W_e=0$ and $\b_e=\mathbf{1}$.
When the embedding mapping does not contain a bias term, it is still possible if the model can figure out some linear combination of features that effectively renders a constant vector. As long as $\x\approx \mathbf{1}$, we will have:
\begin{align}
    f(\bm{x}_m, \bm{x}_n, m-n) &= \bm{x}_m^T\bm{W}_q^T\mathbf{R}_{m-n}^T \bm{W}_k \bm{x}_n\\
    &\approx \sum_{i=1}^{D/2} \cos[(m-n)\theta_i]
    \label{eq:token-rope2}
\end{align}
which also approaches $\delta(m-n)$ as $D\to \infty$.

\subsection{Token-dependent position comparison}
\begin{figure}[h]
    \centering
    \includegraphics[width=0.95\textwidth]{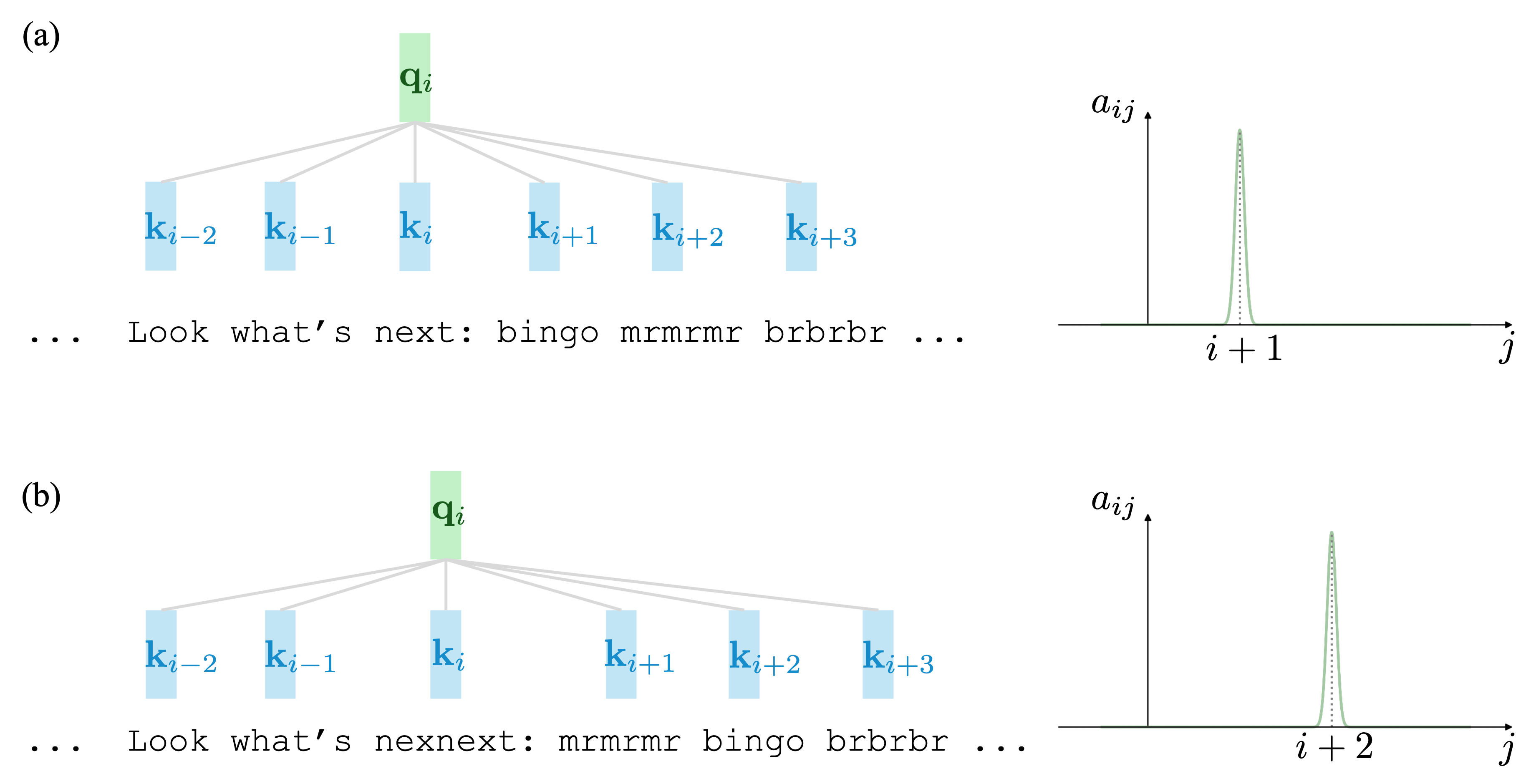}
    \caption{Illustrative task of token-dependent position attending, with the input shown on the left and desired attention weights on the right. Depending on the token value, the desired attention weights focus on the $i+1$-th token (a) and the $i+2$-th token, respectively.}
    \label{fig:toyproblem}
\end{figure}

One key mechanism attention needs to have is to blend the information of token and position. The most basic task is to have a token-dependent position comparison. For example, consider the case illustrated in Fig.~\ref{fig:toyproblem}, where the text input is shown on the left and the ideal attention weights are shown on the right. 
At the $i$-th token, the ideal attention weight depends on the token value. 
If the token is "next", then we need the attention weights to focus on position $i+1$.
If the token is "nexnext", then we need the attention weights to focus on position $i+2$.
The scenario illustrated here is simplistic, but it can be easily generalized to other scenarios, e.g. when the input embedding is in some abstract space instead of simple words, or when causal masks are in place.
The key here is to make sure that each head has the capability to interact the token information with the positional information, which is the cornerstone for a stacked model to function. 

In absolute positional embedding, the functional parameter setting cannot be easily prescribed manually. 
First and foremost, this task definition relies on relative position, which is already hard for absolute positional embedding. While it is possible to prescribe a set of weights that function for a specific position $i$, it is not possible to achieve this functionality for $\forall i\in \mathbb{Z}^+$.
Furthermore, the relative weighting between token information and positional information is fixed in the input embedding, which makes the task impossible for a single layer of attention. Namely, we need:
\begin{align}
    (\x_i + \p_i)^T \W_q^T \W_k (\x_{i+1} + \p_{i+1}) \gg (\x_i + \p_i)^T \W_q^T \W_k (\x_{j} + \p_{j}), \quad \forall j\neq i+1
\end{align}
Note that the values of $\x_{i+1}$ and $\x_{j}$ can be arbitrary, which requires
\begin{align}
    \W_k (\x_j + \p_j) \approx \W_k \p_j
\end{align}
whereas we also need:
\begin{align}
    \W_q (\x_i + \p_i) \approx \W_q \x_i
\end{align}
This creates a tension between the weighting of token information and the weighting of position information in the embedding.

In contrast, RoPE can easily achieve this task within one layer of attention.
In complex notation, we simply need:
{\small
\begin{align}
    \qt^{(n)}_i = \begin{bmatrix}
        e^{-i\theta_1}\\
        e^{-i\theta_2}\\
        \vdots\\
        e^{-i\theta_{D/2}}
    \end{bmatrix},\quad
    \qt^{(nn)}_i = \begin{bmatrix}
        e^{-2i\theta_1}\\
        e^{-2i\theta_2}\\
        \vdots\\
        e^{-2i\theta_{D/2}}
    \end{bmatrix},\quad
    \kt^{(n)}_j = \kt^{(nn)}_j = \begin{bmatrix}
        1\\
        1\\
        \vdots\\
        1
    \end{bmatrix} \forall j
\end{align}
}
where the superscript $(n)$ corresponds to the case where the token is "next" in Fig.~
\ref{fig:toyproblem}(a), and the superscript $(nn)$ corresponds to the case where the token is "nexnext" in Fig.~\ref{fig:toyproblem}(b).
Then the attention is exactly as desired:
{\small
\begin{align}
    \lim_{D\to \infty} \qt_m^{(n)} \Rt^*_{m-n} \kt_n &= \delta(m+1-n)\notag\\
    \lim_{D\to \infty} \qt_m^{(nn)} \Rt^*_{m-n} \kt_n &= \delta(m+2-n)
    \label{eq:delta-rope}
\end{align}
}
This is equivalent to the real form where $\q_i^{(n)}\in\mathbb{R}^D$ has $\q^{(n)}_{i,2t}=\cos(\theta_t)$ and $\q^{(n)}_{i,2t+1}=-\sin(\theta_t)$, while $\q_i^{(nn)}\in\mathbb{R}^D$ has $\q^{(nn)}_{i,2t}=\cos(2\theta_t)$ and $\q^{(nn)}_{i,2t+1}=-\sin(2\theta_t)$.
That can easily result from a choice of projection matrix as detailed in the appendix.
% That can easily result from a choice of projection matrix as in $\qt_i = \Wt_q \xt_i$, where $\xt_i$ is the embedding of the $i$-th token in complex form.
% {\small
% \begin{align}
%     \Wt_q = \begin{bmatrix}
%         e^{-i\theta_1} & e^{-2i\theta_1} & \dots & e^{-i\theta_1} & e^{-2i\theta_1}\\
%         e^{-i\theta_2} & e^{-2i\theta_2} & \dots & e^{-i\theta_2} & e^{-2i\theta_2}\\
%         \vdots & \vdots & \ddots & \vdots & \vdots\\
%         e^{-i\theta_{D/2}} & e^{-2i\theta_{D/2}} & \dots & e^{-i\theta_{D/2}} & e^{-2i\theta_{D/2}}
%     \end{bmatrix},\, 
%     \xt^{(n)}_i = \begin{bmatrix}
%         a_1\\0\\a_2\\0\\ \vdots \\a_{D/4} \\ 0
%     \end{bmatrix}, \,
%     \xt^{(nn)}_i = \begin{bmatrix}
%         0\\a'_1\\0\\a'_2\\0\\ \vdots \\ 0 \\a'_{D/4}
%     \end{bmatrix}
% \end{align}
% }
% Here $a_t$ and $a_t'$ are the degrees of freedom to encode the token information. As long as $\sum_t a_t = \sum_t a'_t$, we can get Eq.~\ref{eq:delta-rope} to hold.

Recall that in general, $\q$ in RoPE when cast to complex form cannot be expressed as $\qt_i = \Wt_q \xt_i$ because the function space of CRoPE is only half that of RoPE. Here we note that the perfect solution to this illustrative task lies exactly within the CRoPE subspace. Namely, for this particular task, half of the function space of RoPE, as well as half the parameters involved, is indeed redundant.

% With this capability, token information and position information can interact within each attention layer. With the help of the feedforward network that comes later, the information can be processed further. By repeating such transformer blocks, it is conceivable that all desired levels of complexity can be achieved.

While this toy problem is simplistic, it is a minimal example to illustrate the advantage of RoPE over conventional absolute positional embedding. We have shown that the same advantage can be obtained by constraining ourselves to the function subspace of CRoPE instead. Note that this example is only for illustration. How well it can extrapolate to deeper networks may be beyond analytical work and invite for empirical study.

%% file: methods.tex
\section{Experiments}
\label{methods}

\subsection{Model Architecture}

We use a decoder-only Transformer with pre-norm RMSNorm ($\epsilon=10^{-6}$) and RoPE ($\theta=5000$). Unless noted otherwise, all models use 16 layers, 8 attention heads, hidden size $d_{\text{model}}=1024$, and a SwiGLU FFN with intermediate size $d_{\text{ff}}=1024$. We apply per-head QK-Norm and tie the token embedding and output projection.

\paragraph{CRoPE parameterization.}
The projection matrix $\W$ will have weights tied as following: for all even indices $i, j$,
$\W_{i,j} = \W_{i+1,j+1}, \W_{i+1,j}=-\W_{i,j+1}$.
% A standard attention projection is a real matrix $W\in\mathbb{R}^{D\times D}$. Since RoPE pairs adjacent coordinates as $z_j=x_{2j}+x_{2j+1} \mathrm{i}$, inputs and outputs can be viewed in $\mathbb{C}^{D/2}$. CRoPE constrains each $2\times2$ block of $W$ to
% \begin{equation}
% W_{jk}=
% \begin{bmatrix}
% A_{jk} & B_{jk} \\
% -B_{jk} & A_{jk}
% \end{bmatrix},
% \end{equation}
% the real form of complex multiplication. Equivalently, with $W_A,W_B\in\mathbb{R}^{\frac{D}{2}\times\frac{D}{2}}$,
% \begin{equation}
% \tilde{y}=(W_A+iW_B)\tilde{x},
% \end{equation}
% and $y$ is formed by interleaving the real and imaginary parts of $\tilde{y}$. Each constrained projection uses $D^2/2$ parameters, a 50\% reduction from $D^2$.

We implement CRoPE with \texttt{BlockLinear}. With \texttt{tied=True}, it enforces the structure above; with \texttt{tied=False}, each $2\times2$ block has four independent entries and is equivalent to a standard linear layer. All computation remains real-valued.

\begin{figure*}[t]
\centering
\begin{minipage}[t]{0.48\textwidth}
\vspace{15pt}
\centering
\scriptsize
\setlength{\tabcolsep}{3pt}
\begin{tabular}{lllc}
\toprule
\textbf{Mode} & \textbf{Q, K} & \textbf{V, Out} & \textbf{Savings} \\
\midrule
\texttt{none}  & Untied & Untied & 0\% \\
\texttt{qk}    & Tied   & Untied & 25\% \\
\texttt{qkv}   & Tied   & Untied (out only) & 37.5\% \\
\texttt{all}   & Tied   & Tied   & 50\% \\
\bottomrule
\end{tabular}
\end{minipage}\hfill
\begin{minipage}[t]{0.5\textwidth}
\vspace{0pt}
\centering
\includegraphics[width=\linewidth]{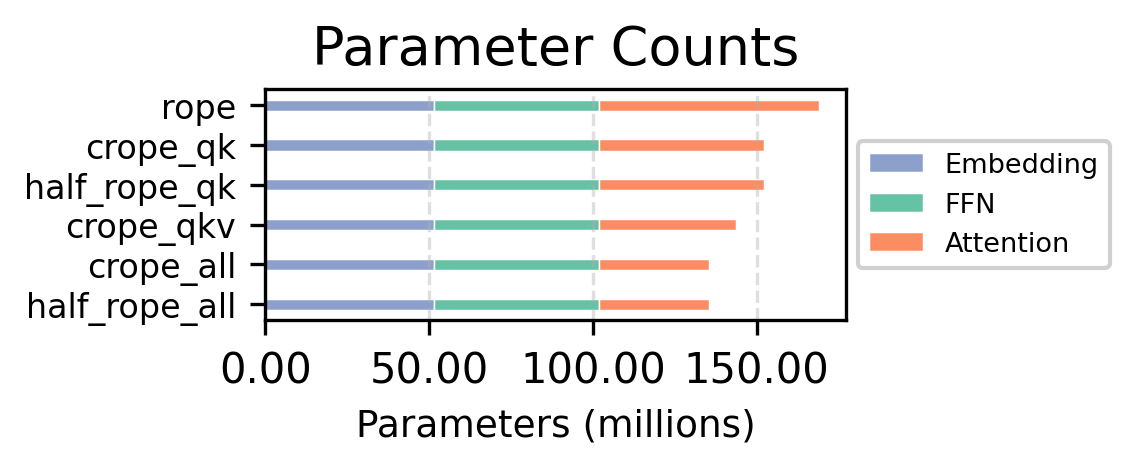}
\end{minipage}
\caption{Left: CRoPE placement modes and theoretical attention-parameter savings. Right: Total parameter counts for each RoPE/CRoPE variant. Embedding and FFN parameters are identical across configurations, so all reductions come from attention.}
\label{fig:param_counts}
\end{figure*}

\paragraph{CRoPE variants.}
We vary where CRoPE is applied in attention, as shown in Figure \ref{fig:param_counts} (left).

We also include parameter-matched dense baselines: \texttt{half\_rope\_qk}, which halves dense Q and K to $d_{\text{model}}/2$, and \texttt{half\_rope\_all}, which additionally halves V and the output projection.

\subsection{Dataset}

We pre-train on FineWeb-Edu dataset \citep{lozhkov2024fineweb-edu} with sequence length of 512.

In addition to the validation loss on FineWeb-Edu dataset, we evaluate zero-shot performance on three benchmarks through \texttt{lm\_eval}~\citep{eval-harness}: HellaSwag~\citep{zellers2019hellaswagmachinereallyfinish}, MMLU~\citep{hendrycks2021measuringmassivemultitasklanguage}, and GSM8K~\citep{cobbe2021trainingverifierssolvemath}.

\subsection{Training Settings}
All runs use Muon~\citep{jordan2024muon} with weight decay $0.1$. Details in appendix.
% We apply a 50-step linear warmup, then cosine decay from $2 \times 10^{-3}$ to $4 \times 10^{-4}$. The batch size is 64 sequences, and each run is trained for 30{,}000 steps. Training uses \texttt{bfloat16} autocast with distributed data-parallel execution launched via \texttt{torchrun}.

%% file: results.tex
\section{Results}
\label{sec:results}

\subsection{Training Dynamics and Parameter Efficiency}
\label{sec:loss_dynamics}

We first verify that the tested variants realize the intended parameter counts. As shown in Figure~\ref{fig:param_counts} (right), embedding and FFN parameters remain fixed across models, while attention parameters decrease according to the applied CRoPE constraint. In particular, the measured reductions for \texttt{crope\_qk}, \texttt{crope\_qkv}, and \texttt{crope\_all} exactly match the expected attention-parameter savings of 25\%, 37.5\%, and 50\%, respectively. \texttt{crope\_qk} and   \texttt{crope\_all} share same parameter counts with \texttt{half\_rope\_qk} and \texttt{half\_rope\_all}, as expected.

\begin{figure*}[th]
\centering
\begin{minipage}[t]{0.49\textwidth}
\centering
\includegraphics[width=\linewidth]{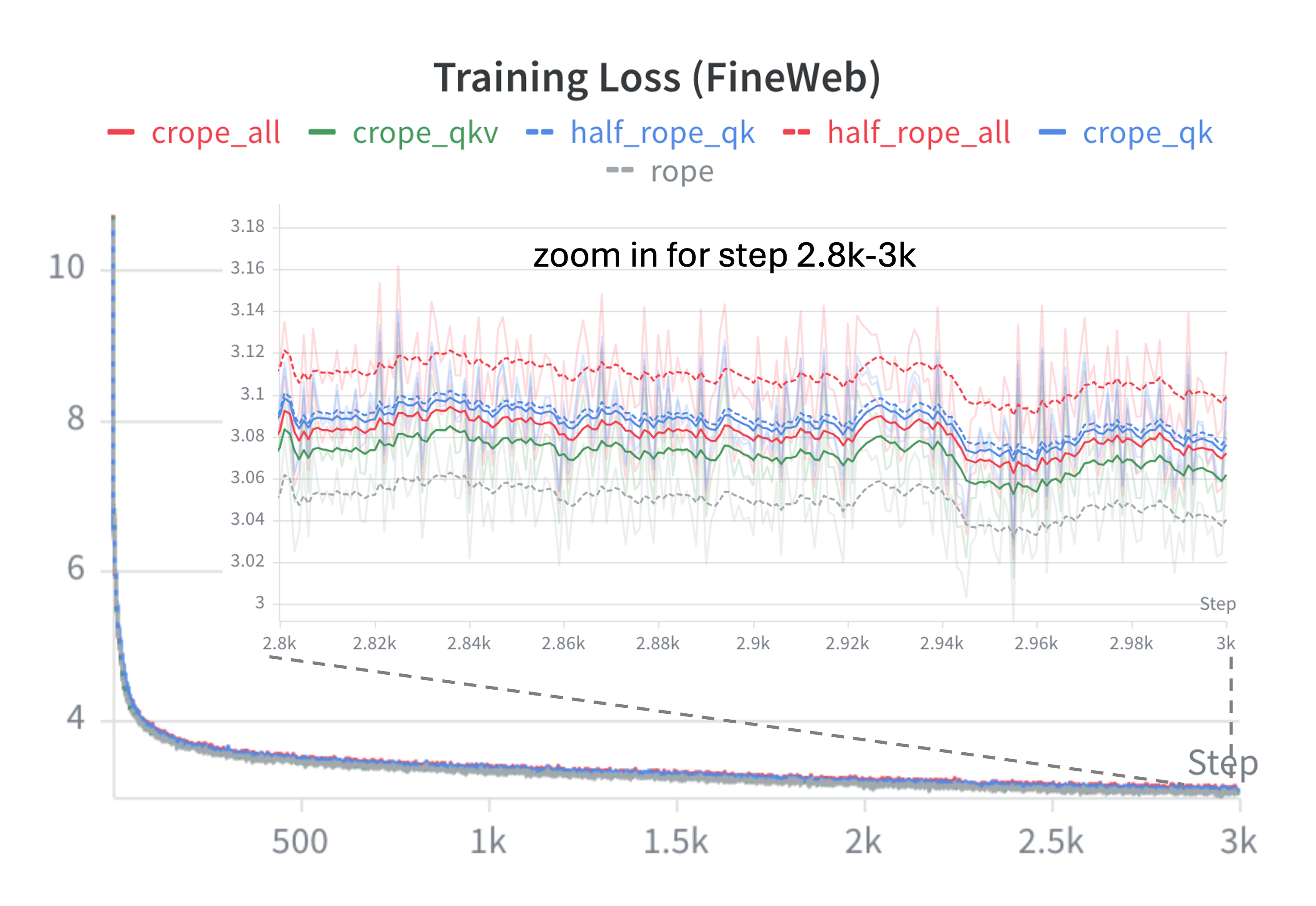}
\end{minipage}\hfill
\begin{minipage}[t]{0.49\textwidth}
\centering
\includegraphics[width=\linewidth]{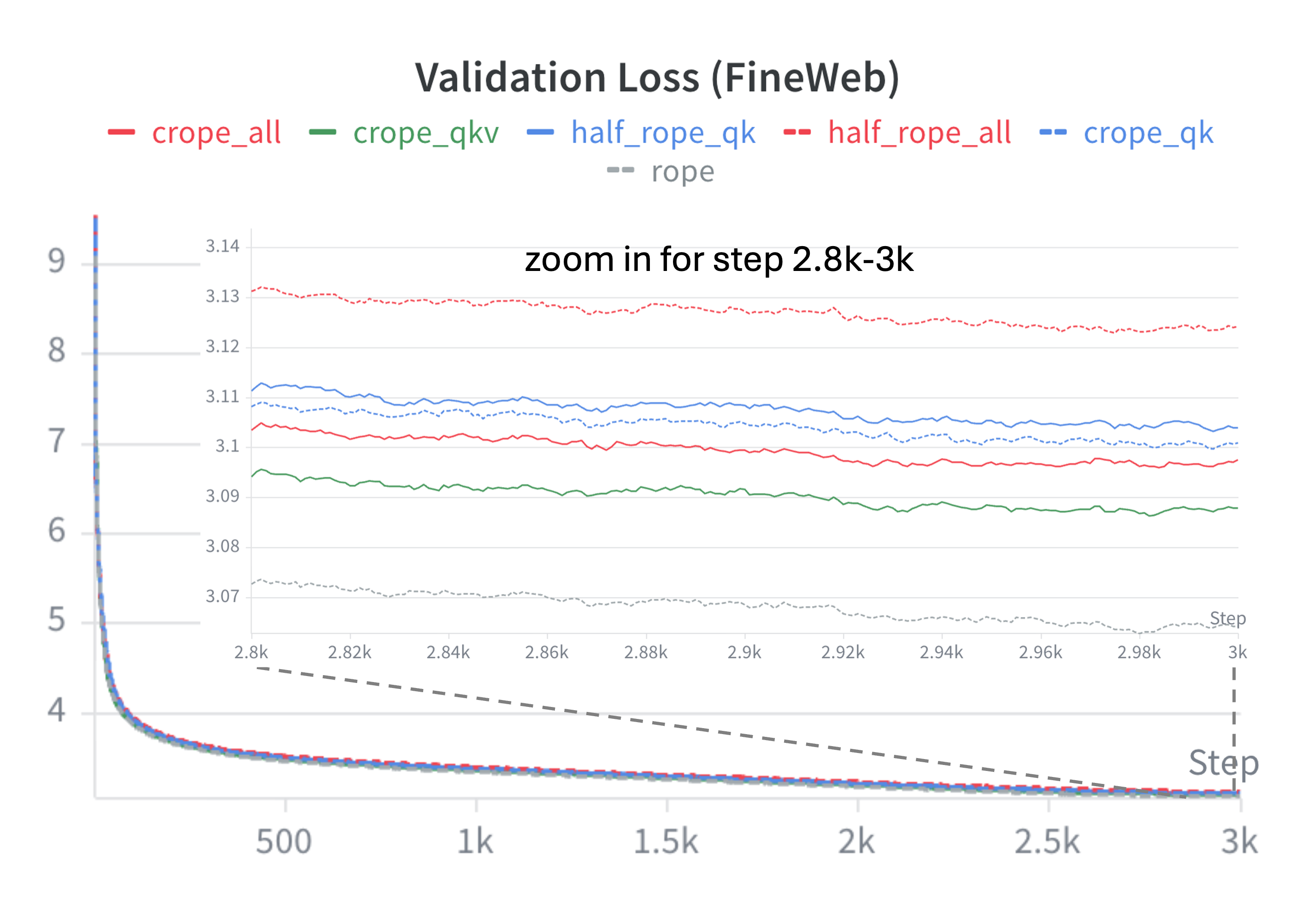}
\end{minipage}
\caption{Training loss (left) and validation loss (right) over optimization steps for all variants. Insets zoom into the final 200 steps to highlight the final performance ordering, training-loss insets are smoothed for readability. Solid lines denote CRoPE variants, dashed lines denote standard or half-width RoPE baselines, and parameter-matched models share the same color.}
\label{fig:train_val_loss}
\end{figure*}

Given these parameter-matched comparisons, Figure~\ref{fig:train_val_loss} shows that all variants optimize stably under the same training schedule. The full-capacity \texttt{rope} baseline achieves the lowest validation loss overall, as expected. More importantly, among reduced-parameter models, CRoPE consistently outperforms its dense half-width counterparts: \texttt{crope\_all} converges below \texttt{half\_rope\_all}, and \texttt{crope\_qk} likewise outperforms \texttt{half\_rope\_qk}. \texttt{crope\_qkv} even outperforms \texttt{half\_rope\_qk} with less parameters. This suggests that the gains are not solely due to parameter count, but also to the complex-linear structure, which yields a more parameter-efficient attention parameterization than simply shrinking dense projections.

\subsection{Downstream Evaluation}
\label{sec:downstream_eval}

\begin{figure*}[t]
\centering
\includegraphics[width=\textwidth]{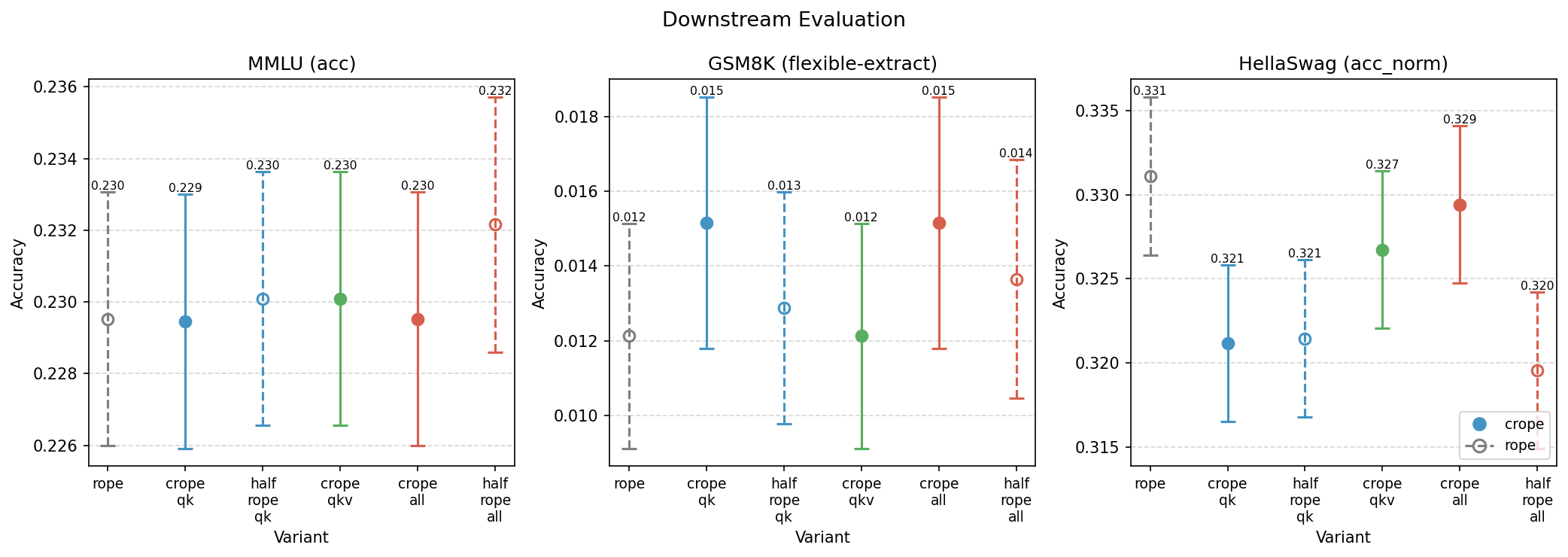}
\caption{Zero-shot downstream evaluation on MMLU (accuracy), GSM8K (flexible-extract), and HellaSwag (normalized accuracy). Error bars denote standard error. Colors and markers follow the parameter-matching convention from Figure~\ref{fig:train_val_loss}.}
\label{fig:downstream}
\end{figure*}

To test whether the validation-loss trends translate to downstream capability, we evaluate all models zero-shot on HellaSwag, MMLU, and GSM8K (Figure~\ref{fig:downstream}).

On HellaSwag, the full \texttt{rope} baseline performs best, reaching 0.331 normalized accuracy. Importantly, CRoPE nearly matches this result at 0.329, with \texttt{crope\_all} close behind at 0.327. Both substantially outperform the parameter-matched dense baseline \texttt{half\_rope\_all}, which achieves 0.320.

In contrast, results on MMLU and GSM8K are tightly clustered across all variants. MMLU accuracy falls in a narrow range from 0.229 to 0.232, with overlapping standard errors indicating no clear difference between configurations. GSM8K flexible-extract scores are likewise uniformly low, ranging from 0.012 to 0.015. This limited spread is expected at this model scale, where zero-shot mathematical reasoning remains weak.

Overall, the downstream results are consistent with the validation-loss trends. CRoPE preserves task performance while substantially reducing attention parameters, and it compares favorably to naive dense width reduction, especially on tasks such as HellaSwag where the baseline model exhibits meaningful signal.